\def\year{2020}\relax
\documentclass[letterpaper]{article} % DO NOT CHANGE THIS
\usepackage{aaai20}  % DO NOT CHANGE THIS
\usepackage{times}  % DO NOT CHANGE THIS
\usepackage{helvet} % DO NOT CHANGE THIS
\usepackage{courier}  % DO NOT CHANGE THIS
\usepackage[hyphens]{url}  % DO NOT CHANGE THIS
\usepackage{graphicx} % DO NOT CHANGE THIS
\usepackage{graphicx}  % DO NOT CHANGE THIS
\usepackage{amsmath}
\usepackage{amsfonts}
\usepackage{algorithm, algorithmic}
\usepackage{siunitx}
\usepackage{enumitem}
\usepackage[T1]{fontenc}

\newcommand{\eat}[1]{}

\newcommand{\E}{\mathbb{E}}

\title{Detecting and Tracking Communal Bird Roosts in Weather Radar Data}
\author{
Zezhou Cheng \\
UMass Amherst \\
\texttt{\small zezhoucheng@cs.umass.edu}
\And
Saadia Gabriel \\
University of Washington \\
{\tt\small skgabrie@cs.washington.edu}
\And
Pankaj Bhambhani \\
UMass Amherst \\
{\tt \small pankaj@cs.umass.edu} \\
\AND
Daniel Sheldon \\
UMass Amherst \\
{\tt \small sheldon@cs.umass.edu}
\And 
Subhransu Maji \\
UMass Amherst \\
{\tt \small smaji@cs.umass.edu}
\And
Andrew Laughlin \\
UNC Asheville \\
{\tt \small alaughli@unca.edu}
\And 
David Winkler \\
Cornell University \\
{\tt\small  dww4@cornell.edu}
}

\begin{document}

\maketitle

\begin{abstract}
    The US weather radar archive holds detailed information about
    biological phenomena in the atmosphere over the last 20
    years.
    Communally roosting birds congregate in large numbers at
    nighttime roosting locations, and their morning exodus from the
    roost is often visible as a distinctive pattern in radar
    images. This paper describes a machine learning system to detect
    and track roost signatures in weather radar data. A significant
    challenge is that labels were collected opportunistically from
    previous research studies and there are systematic differences in
    labeling style. We contribute a latent-variable model and
    EM algorithm to learn a detection model together with models of
    labeling styles for individual annotators.
    By properly accounting for these variations we learn a
    significantly more accurate detector. The resulting system detects
    previously unknown roosting locations and provides comprehensive
    spatio-temporal data about roosts across the US. This data will
    provide  biologists important information about the poorly understood phenomena of
    broad-scale habitat use and movements of communally roosting birds
    during the non-breeding season.
\end{abstract}

\section{Introduction}
\label{s:intro}

The US weather radar network offers an unprecedented opportunity to study dynamic, continent-scale movements of animals over a long time period. The National Weather Service operates a network of 143 radars in the contiguous US. These radars were designed to study weather phenomena such as precipitation and severe storms. However, they are very sensitive and also detect flying animals, including birds, bats, and insects~\cite{kunz2008aeroecology}. The data are archived and available from the early 1990s to present, and provide comprehensive views of a number of significant biological phenomena. These include broad-scale patterns of bird migration, such as the density and velocity of all nocturnally migrating birds flying over the US in the spring and fall~\cite{Farnsworth2016,shamoun2016innovative}, and the number of migrants leaving different stopover habitats~\cite{buler2009quantifying,buler2014radar}. Radars can also detect phenomena that can be matched to individual species, such as insect hatches and departures of large flocks of birds or bats from roosting locations~\cite{winkler2006roosts,horn2008analyzing,buler2012mapping}.

\begin{figure*}[ht]
  \includegraphics[width=\linewidth]{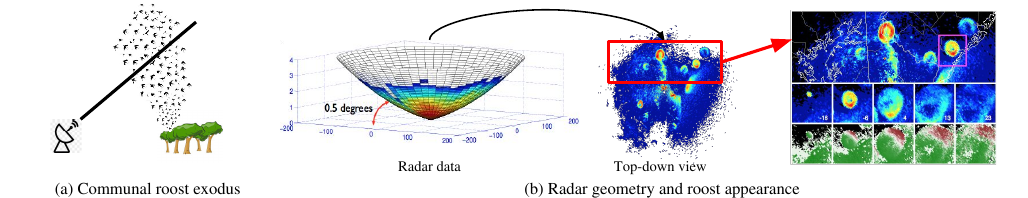}
  \caption{\label{fig:roost-example}
    (a) Illustration of roost exodus.
    (b) A radar traces out cone-shaped slices of the atmosphere (left),
    which are rendered as top-down images (center). This image from
    the Dover, DE radar station at 6:52 am on Oct 2, 2010 shows at
    least 8 roosts. Several are shown in more detail to the right, together with
    crops of one roost from five consecutive reflectivity and radial
    velocity images over a period of 39 minutes. These show the distinctive expanding ring and 
    ``red-white-green'' diverging velocity patterns.}
\end{figure*}

This paper is about the development of an AI system to track the roosting and migration behavior of communally roosting songbirds.  Such birds congregate in large roosts at night during portions of the year.  Roosts may contain thousands to millions of birds and are packed so densely that when birds depart in the morning they are visible on radar. Roosts occur all over the US, but few are documented, and many or most are never witnessed by humans, since birds enter and leave the roosts during twilight. Those who do witness the swarming behavior of huge flocks at roosts report an awe-inspiring spectacle\footnote{See: \url{https://people.cs.umass.edu/~zezhoucheng/roosts}}.

We are particularly interested in swallow species, which, due to their unique behavior, create a distinctive expanding ring pattern in radar (see Fig.~\ref{fig:roost-example}). During fall and winter, the Tree Swallow (\emph{Tachycineta bicolor}) is the only swallow species present in North America, so roost signatures provide nearly unambiguous information about this \emph{single} species, which is highly significant from an ecological perspective.  Swallows are aerial insectivores, which are rapidly declining in North America~\cite{Fraser2012,Nebel2010}. Information about the full life cycles of these birds is urgently needed to understand and reverse the causes of their declines~\cite{north2012state}. Radar data represents an unparalleled opportunity to gather this information, but it is too difficult to access manually, which has greatly limited the scope (in terms of spatial/temporal extent or detail) of past research studies~\cite{Laughlin2016,bridge2016}. 

We seek an AI system to fill this gap. An AI system has the potential to collect fine-grained measurements of swallows---across the continent, at a daily time scale---from the entire 20-year radar archive. This would allow scientists to study trends in population size and migration behavior in relation to changes in climate and habitat availability, and may provide some of the first insights into where and when birds die during their annual life cycle, which is critical information to guide conservation efforts.

The dramatic successes of deep learning for recognition tasks make it an excellent candidate for detecting and tracking roosts in radar. We develop a processing pipeline to extract useful biological information by solving several challenging sub-tasks. First, we develop a single-frame roost detector system based on Faster R-CNN, an established object detection framework for natural images~\cite{ren2015faster}. Then, we develop a tracking system based on the ``detect-then-track'' paradigm~\cite{ren2008finding} to assemble roosts into sequences to compute meaningful biological measures and improve detector performance. Finally, we use auxiliary information about precipitation and wind farms to reduce certain sources of false positives that are poorly represented in our training data. In the final system, 93.5\% of the top 521 tracks on fresh test data are correctly identified and tracked bird roosts.

A significant challenge in our application is the presence of systematic differences in labeling style. Our training data was annotated by different researchers and naturalists using a public tool developed for prior research studies~\cite{Laughlin2016}. They had different backgrounds and goals, and roost appearance in radar is poorly understood, leading to considerable variability, \emph{much of which is specific to individual users}. This variation makes evaluation using held out data very difficult and inhibits learning due to inconsistent supervision.

We contribute a novel approach to jointly learn a detector together with user-specific models of labeling style. We model the true label $y$ as a latent variable, and introduce a probabilistic \emph{user model} for the observed label $\hat{y}$ conditioned on the image, true label, and user.  We present a variational EM learning algorithm  that permits learning with only black-box access to an existing object detection model, such as Faster R-CNN. We show that accounting for user-specific labeling bias significantly improves evaluation and learning.

Finally, we conduct a case study by applying our models to detect roosts across the entire eastern US on a daily basis for fall and winter of 2013-2014 and cross-referencing roost locations with habitat maps. The detector has excellent precision, detects previously unknown roosts, and demonstrates the ability to generate urgently needed continent-scale information about the non-breeding season movements and habitat use of an aerial insectivore. Our data points to the importance of the eastern seaboard and Mississippi valley as migration corridors, and suggests that birds rely heavily on croplands (e.g., corn and sugar cane) earlier in the fall prior to harvest.

\newcommand{\new}[1]{#1}
\section{A System to Detect and Track Roosts}\label{s:application}

\paragraph{Radar Data}
We use radar data from the US NEXRAD network of over 140
radars operated by the National Weather Service~\cite{crum1993wsr}.
They have ranges of several hundred kilometers and cover
nearly the entire US.
Data is available from the 1990s to present in the form of
raster data products summarizing the results of radar \emph{volume
scans}, during which a radar scans the surrounding airspace by
rotating the antenna 360$^\circ$ at different elevation angles (e.g., 0.5$^\circ$, 1.5$^\circ$)
to sample a cone-shaped ``slice'' of airspace~(Fig.~\ref{fig:roost-example}b).
Radar scans are available every 4--10 minutes at each station.
Conventional radar images are top-down views of these
sweeps; we will also render data
this way for processing.

Standard radar scans collect 3 data products at 5 elevation angles, for 15 total channels.
We focus on data products that are most relevant for detecting
roosts.
\emph{Reflectivity} is the base measurement of the density of objects in the
atmosphere.
\emph{Radial velocity} uses the Doppler shift of the returned signal
to measure the speed at which objects are approaching or departing the
radar.
\emph{Copolar cross-correlation coefficient} is a newer data product, available
since 2013, that is useful for discriminating rain from
biology~\cite{stepanian2016dual}. We use it for post-processing,
but not training, since most of our labels are from before 2013.

\paragraph{Roosts}
A roost exodus (Fig.~\ref{fig:roost-example}a) is the mass departure
of a large flock of birds from a nighttime roosting location. 
They occur 15--30 minutes before sunrise and are very rarely witnessed
by humans.
However, roost signatures are visible on radar as birds fly
upward and outward into the radar domain.
Fig.~\ref{fig:roost-example}b, center, shows a radar reflectivity
image with at least 8 roost signatures in a \SI{300 x 300}{km}
area.
Swallow roosts, in particular, have a characteristic signature shown
in Fig.~\ref{fig:roost-example}b, right.
The center row shows reflectivity images of one roost expanding over
time.
The bottom row shows the characteristic radial velocity pattern of
birds dispersing away from the center of the roost.
Birds moving toward the radar station (bottom left) have negative
radial velocity (green) and birds moving away from the radar station
(top right) have positive radial velocity (red).

\paragraph{Annotations}
We obtained a data set of manually annotated
roosts collected for prior ecological
research~\cite{Laughlin2016}.
They are believed to be nearly 100\% Tree Swallow roosts.
Each label records the position and
radius of a circle within a radar image that best approximates the
roost. We restricted to seven stations in the eastern US and to
month-long periods that were exhaustively labeled, so we could infer
the absence of roosts in scans with no labels. 
We restricted to scans from 30 minutes before to 90 minutes after
sunrise, leading to a data set of 63691 labeled roosts in 88972 radar
scans.
A significant issue with this data set is systematic
differences in labeling style by different researchers. This poses
serious challenges to building and evaluating a detection model. 
We discuss this further and present a methodological solution in
Sec.~\ref{s:approach}.

\paragraph{Related Work on Roost Detection and Tracking}

There is a long history to the study of roosting behavior with the radar data, almost entirely based on human interpretation of images~\cite{winkler2006roosts,Laughlin2016,bridge2016}. That work is therefore restricted to analyze only limited regions, short-time periods or coarse-grained information about the roosts. \citeauthor{chilson2019}~[\citeyear{chilson2019}] developed a deep-learning image classifier to identify radar images that contain roosts. While useful, this provides only limited biological information. Our method locates roosts within images and tracks them across frames, which is a substantially harder problem, and important biologically. For example, to create population estimates or locate roost locations on the ground, one needs to know where roosts occur within the radar image; to study bird behavior, one needs to measure how roosts move and expand from frame to frame. Our work is the first machine-learning method able to extract the type of higher-level biological information presented in our case study (Sec.~\ref{s:study}). Our case study illustrates, to the best of our knowledge, the \emph{first} continent-scale remotely sensed observations of the migration of a single bird species.

\subsection{Methodology}

\begin{figure}
\includegraphics[width=\linewidth]{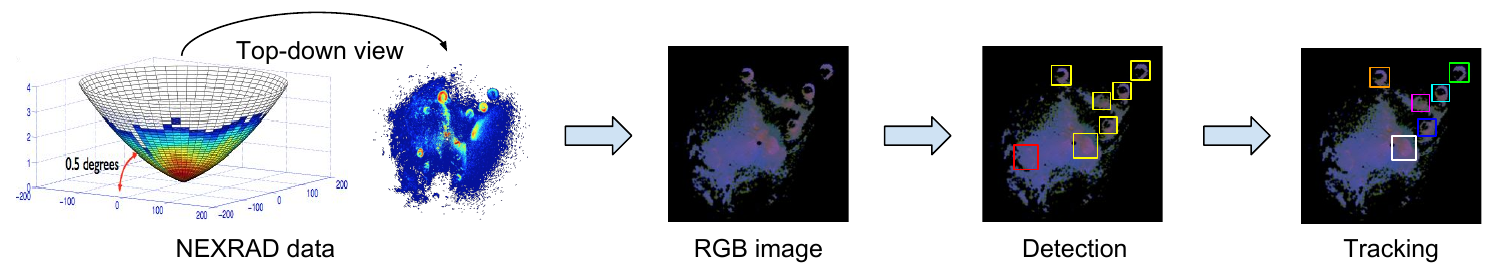}
\caption{Detection and tracking pipeline. A final step (not shown) uses
  auxiliary data to filter rain and wind-farms.}
\label{fig:pipeline}
\end{figure}

Our overall approach consists of four steps (see
Fig.~\ref{fig:pipeline}): we render radar scans as
multi-channel images, run a single-frame detector, assemble and
rescore tracks, and then postprocess detections using other geospatial
data to filter specific sources of false positives.

\paragraph{Detection Architecture}
Our single-frame detector is based on Faster R-CNNs~\cite{ren2015faster}.
Region-based CNN detectors such as Faster R-CNNs are 
state-of-the-art on several object detection benchmarks.

A significant advantage of these architectures comes from pretraining parts of the network on large labeled image datasets such as ImageNet~\cite{deng2009imagenet}.
To make radar data compatible with these networks, we must select only 3 of the 15 available channels to feed into the RGB-based models.
We select the radar products that are most discriminative for humans: reflectivity at 0.5$^\circ$, radial velocity at 0.5$^\circ$ degrees, and reflectivity at 1.5$^\circ$. Roosts appear predominantly in the lowest elevations and are distinguished by the ring pattern in reflectivity images and distinctive velocity pattern. 
These three data products are then rendered as a $1200\times1200$ image in the
``top-down'' Cartesian-coordinate view (out to 150km from the radar
station) resulting in a 3-channel $1200\times1200$ image. 
The three channel images are fed into Faster R-CNN
initialized with a pretrained VGG-M network~\cite{chatfield2014return}.
All detectors are trained for the single ``roost'' object class, using
bounding boxes derived from the labeled dataset described above.

Although radar data is visually different from natural images, we found
ImageNet pretraining to be quite useful; without pretraining the networks
 took significantly longer to converge and resulted in a 15\% lower
 performance. 
We also experimented with models that map 15 radar channels down to 3
using a learned transformation. 
These networks were \emph{not consistently better} than ones using hand-selected channels.
 Models trained with shallower networks that mimic handcrafted
 features, 
such as those based on gradient histograms, %~\cite{dalal2005histograms}
performed 15-20\% worse depending on the architecture. See
the \emph{supplementary material} on the project page for details on these baseline detection models.

We defer training details to Sec.~\ref{s:approach}, where we discuss our approach of jointly learning the Faster R-CNN detector together with user-models for labeling style.

\paragraph{Roost Tracking and Rescoring}\label{Sec:tracking}
Associating and tracking detections across frames is important for several reasons. It helps rule out false detections due to rain and other phenomenon that have different temporal properties than roosts (see Sec.~\ref{s:study}).
Detection tracks are also what associate directly to the biological entity---a single flock of birds---so they are needed to estimate biological parameters such as roost size, rate of expansion, location and habitat of first appearance, etc.
We employ a greedy heuristic to assemble detections from individual frames into tracks~\cite{ren2008finding}, starting with high scoring detections and incrementally adding unmatched detections with high overlap in nearby frames.
Detections that match multiple tracks are assigned to the longest one.
After associating detections we apply a Kalman smoother to each track
using a linear dynamical system model for the bounding box center and
radius. This model captures the dynamics of roost formation and growth
with parameters estimated from ground-truth annotations.
We then conduct a final rescoring step where track-level features
(e.g., number of frames, average detection score of all bounding boxes
in track) are associated to individual detections, which are then
rescored using a linear SVM.
This step suppresses false positives that appear roost-like in single frames but do not behave like roosts.
Further details of association, tracking and rescoring can be found in the supplementary material.

\paragraph{Postprocessing with Auxiliary Information} \label{Sec: dual-pol}
In preliminary experiments, the majority of high-scoring tracks were roosts, but there were also a significant number of high-scoring false positives caused by specific phenomena, especially wind farms and precipitation (see Sec.~\ref{s:study}).
We found it was possible to reliably reject these false positives using auxiliary information. 
To eliminate rain in modern data, we use the radar measurement of copolar cross-correlation coefficient,
$\rho_{HV}$, which is available since 2013~\cite{stepanian2016dual}.
Biological targets have much lower $\rho_{HV}$ values than
precipitation due to their high variance in orientation, position and
shape over time. A common rule is to classify pixels as rain if
$\rho_{HV} > 0.95$~\cite{dokter2018biorad}. We classify a roost
detection as precipitation if a majority of pixels inside its bounding
box have $\rho_{HV}>0.95$.
For historic data one may use automatic methods for segmenting
precipitation in radar images such as~\cite{mistnet}.
For wind farms, we can use recorded turbine locations from the
U.S. Wind Turbine Database~\cite{USWTDB}. A detection is identified as
a wind farm if any turbine from the database is located inside its
bounding box.

\section{Modeling Labeling Styles}\label{s:approach}

\begin{figure}
  \includegraphics[width=\columnwidth]{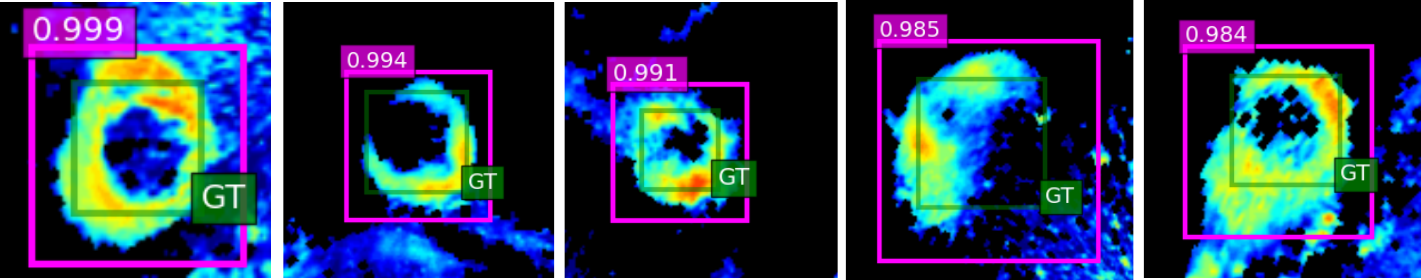}
  \caption{\label{fig:noisy-annotations}Labeling style variation leads to inaccurate evaluation and suboptimal detectors. All of these detections (pink boxes) are misidentified as false positives because of insufficient overlap with annotations of one user (green boxes) with a tight labeling style. Label variation also hurts training and leads to suboptimal models.}
\end{figure}

Preliminary experiments revealed that systematic variations in labeling style were a significant barrier to training and evaluating a detector.
Fig.~\ref{fig:noisy-annotations} shows example detections that correctly locate and circumscribe roosts, but are classified as false positives because the annotator used labels (originally circles) to ``trace'' roosts instead of circumscribing them.
Although it is clear upon inspection that these detections are ``correct'', with 63691 labels and a range of labeling styles, there is no simple adjustment to accurately judge performance of a system.
Furthermore, labeling variation also inhibits learning and leads to suboptimal models.
This motivates our approach to jointly learn a detector along with user-specific models of labeling style.
Our goal is a generic and principled approach that can leverage existing detection frameworks with little or no modification.

\paragraph{Latent Variable Model and Variational EM Algorithm} 
To model variability due to annotation styles we use the following graphical model:
\begin{center}
  \includegraphics[width=0.7\columnwidth]{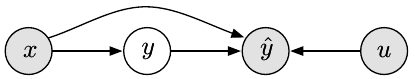}
\end{center}
where $x$ is the image, $y$ represents the unobserved ``true'' or gold-standard label, $u$ is the user (or features thereof), and $\hat{y}$ is the observed label in user $u$'s labeling style. In this model
\begin{itemize}[leftmargin=*, itemsep=0pt]
\item $p_{\theta}(y | x)$ is the \emph{detection model}, with parameters $\theta$.
We generally assume the negative log-likelihood of the detection model is equal to the loss function of the base detector. For example, in our application, $-\log p_{\theta}(y | x) = L_{\text{cnn}}(\theta|y)$, the loss function of Faster R-CNN.\footnote{Faster R-CNN includes a region proposal network to detect and localize candidate objects and a classification network to assign class labels. The networks share parameters and are trained jointly to minimize a sum of several loss functions; we take the set of all parameters as $\theta$ and the sum of loss functions as $L_{\text{cnn}}(\theta|y)$.}
  
\item $p_{\beta}(\hat{y} \mid x, y, u)$ is the \emph{forward user model} for the labeling style of user $u$, with parameters $\beta$. In our application, much of the variability can be captured by user-specific scaling of the bounding boxes, so we adopt the following user model: for each bounding-box we model the observed radius as $p_{\beta}(\hat{r} \mid r, u) = \mathcal{N}(\hat{r}; \beta_u r, \sigma^2)$ where $r$ is the unobserved true radius and $\beta_u$ is the user-specific scaling factor. In this model, the bounding-box centers are unmodified and the user model does not depend on the image $x$, even though our more general framework allows both.
 
\item $p_{\theta,\beta}(y \mid x, \hat{y}, u)$ is the \emph{reverse user model}. It is determined by the previous two models, and is needed to reason about the true labels given the noisy ones during training. Since this distribution is generally intractable, we use instead a variational reverse user model $q_\phi(y \mid x, \hat{y}, u)$, with parameters $\phi$. In our application, $q_\phi(r \mid \hat{r}, u) = \mathcal{N}(r; \phi_u \hat{r}, \sigma^2)$, which is another user-specific rescaling of the radius. 

\end{itemize}

We train the user models jointly with Faster R-CNN using variational EM.
We initialize the Faster R-CNN parameters \(\theta\) by training for 50K iterations starting from the ImageNet pretrained VGG-M model using the original uncorrected labels. 
We then initialize the forward user model parameters $\beta$ using the Faster R-CNN predictions: if a predicted roost with radius $r_i$ overlaps sufficiently with a labeled roost (intersection-over-union > 0.2) and has high enough detection score (> 0.9), we generate a training pair $(r_i, \hat{r}_i)$ where $\hat{r}_i$ is the labeled radius. We then estimate the forward regression model parameters as a standard linear regression with these pairs. 

After initialization, we repeat the following steps (in which $i$ is an index for annotations):
\begin{enumerate}[leftmargin=*,itemsep=0pt]
\item Update parameters $\phi$ of the reverse user model by minimizing the combined loss \( \E_{r_i \sim q_\phi(r_i | \hat{r}_i, u_i)} \big[L_{\text{cnn}}(\theta | \{r_i\}) - \sum_i \log p_{\beta}(\hat{r}_i | r_i, u_i) \big]\). 
  The optimization is performed separately to determine the reverse scaling factor $\phi_u$ for each user using Brent's method with search boundary $[0.1, 2]$ and black-box access to \(L_{\text{cnn}}\).
 \item Resample annotations on the training set by sampling $r_i \sim q_\phi(\cdot | \hat{r}_i, u_i)$ for all $i$, then update $\theta$ by training Faster R-CNN for 50K iterations using the resampled annotations.
 \item Update $\beta$ by training the forward user models using pairs $(r_i, \hat{r}_i)$, where $r_i$ is the radius of the imputed label.
\end{enumerate}

Formally, each step can be justified as maximizing the \emph{evidence lower bound} (ELBO)~\cite{blei2017variational} of the log marginal likelihood $\log p_{\theta, \beta}(\hat{y} | x, u) = \log \int p_{\theta,\beta}(\hat{y}, y | x, u) dy$ with respect to the variational distribution $q_\phi$. Steps 1, 2, and 3 maximize the ELBO with respect to $\phi$, $\theta$, and $\beta$, respectively.
Steps 1 and 2 require samples from the reverse user model; we found that using the \emph{maximum a posteriori} $y$ instead of sampling is simple and performs well in practice, so we used this in our application.

The derivation is presented in the supplementary material. It assumes that $y$ is a \emph{structured} label that includes all bounding boxes for an image. This justifies equating $- \log p_\theta(y | x)$ with the loss function $L(\theta)$ of an existing detection framework that predicts bounding boxes simultaneously for an entire image (e.g., using heuristics like non-maximum suppression). This is important because it is modular. We can use any detection framework that provides a loss function, with no other changes.
A typical user model will then act on $y$ (a set of bounding boxes) by acting independently on each of its components, as in our application.

We anticipate this framework can be applied to a range of applications. More sophisticated user models may also depend on the image $x$ to capture different labeling biases, such as different thresholds for labeling objects, or tendencies to mislabel objects of a certain class or appearance. However, it is an open question how to design more complex user models and we caution about the possibility of very complex user models ``explaining away'' true patterns in the data.

\paragraph{Related Work on Labeling Style and Noise}
\citeauthor{jiang2017face}~[\citeyear{jiang2017face}] discuss how systematic differences in labeling style across face-detection benchmarks significantly complicate evaluation, and propose fine-tuning techniques for style adaptation. Our EM approach is a good candidate to unify training and evaluation across these different benchmarks. Prior research on label \emph{noise}~\cite{frenay2014classification} has observed that noisy labels degrade classifier performance~\cite{nettleton2010study} and proposed various methods to deal with noisy labels~\cite{van2015learning,ghosh2017robust,brodley1999identifying,xiao2015learning,tanaka2018joint}, including EM~\cite{mnih2012learning}.  While some considerations are similar (degradation of training performance, latent variable models), labeling \emph{style} is qualitatively different in that an explanatory variable (the user) is available to help model systematic label variation, as opposed to pure ``noise''. Also, the prior work in this literature is for classification. Our approach is the first noise-correction method for bounding boxes in object detection.

\section{Experiments} \label{s:experiments}

\paragraph{Dataset} We divided the 88972 radar scans from the manually labeled dataset (Sec.~\ref{s:application}) into training, validation, and test sets. 
Tab.~\ref{tab_exp_variants} gives details of training and test data by station. The validation set (not shown) is roughly half the size of the test set and was used to set the hyper-parameters of the detector and the tracker.

\paragraph{Evaluation Metric}
To evaluate the detector we use established evaluation metrics for object
detection employed in common computer vision benchmarks.
A detection is a true positive if its overlap with an annotated bounding-box, measured using the intersection-over-union (IoU) metric, is greater than 0.5.
The mean average precision (MAP) is computed as the area under the precision-recall curve.
For the purposes of evaluating the detector we mark roosts smaller than \num{30 x 30} in a \num{1200 x 1200} radar image as difficult and ignore them during evaluation.
Humans typically detect such roosts by looking at adjacent frames.
As discussed previously (Fig.~\ref{fig:noisy-annotations}), evaluation is unreliable when user labels have different labeling styles. To address this we propose an evaluation metric (``+User'') that rescales predictions on a per-user basis prior to computing MAP.
Scaling factors are estimated following the same procedure used to initialize variational EM. This assumes that the user information is known for the test set, where it is \emph{only} used for rescaling predictions, and not by the detector.

\paragraph{Results: Roost Detector and User Model}
Tab.~\ref{tab_exp_variants} shows the performance of various
detectors across radar stations.
We trained two detector variants, one a standard Faster
R-CNN, and another trained with the variational EM algorithm.
We evaluated the detectors based on whether annotation bias was
accounted for during testing (Tab.~\ref{tab_exp_variants}, ``+User'').

The noisy annotations cause inaccurate evaluation.
A large number of the detections on \texttt{KDOX} are misidentified as negatives because of the low overlap with the annotations, which are illustrated in Fig.~\ref{fig:noisy-annotations}, leading to a low MAP score of 9.1\%. 
This improves to 44.8\% when the annotation biases are accounted for during testing.
As a sanity check we trained and evaluated a detector on annotations of a single user on \texttt{KDOX} and found its performance to be in the mid fifties. 
However, the score was low when this model was evaluated on annotations from other users or stations.
 
The detector trained jointly with user-models using variational EM further improves performance across all stations~(Tab.~\ref{tab_exp_variants}, ``+EM+User''), with larger improvements for stations with less training data. Overall MAP improves from 44.2\% to 45.5\%. 
To verify the statistical significance of this result, we drew 20 sets of bootstrap resamples from the entire test set (contains $23.7k$ images), computed the MAP of the model trained with EM and without EM on each set. The mean and standard error of MAPs for the model trained with EM are 45.5\% and 0.12\% respectively, while they are 44.4\% and 0.11\% for the model trained without EM.

\begin{table}[t]
\small
\centering
\begin{tabular}{ c|c|c|c|c|c}
\textbf{Station} &  \textbf{Test} & \textbf{Train} & \textbf{R-CNN} & \textbf{+User} & \textbf{+EM+User} \\
\hline
\texttt{KMLB}  &9133  & 19998 &  47.5 & 47.8 & \textbf{49.2} \\
\texttt{KTBW}  & 7195& 16382 &  47.3 & 50.0 & \textbf{50.8}\\
\texttt{KLIX}  & 4077& 10192 &  32.4 & 35.1 & \textbf{35.7}\\
\texttt{KOKX}  & 1404 & 2994 &  23.2 & 27.3 & \textbf{29.9}\\
\texttt{KAMX}  & 860& 1898 &  29.9 & 30.8 & \textbf{31.6}\\
\texttt{KDOX}  & 639 & 902&  9.1 & 44.8 & \textbf{50.2}\\
\texttt{KLCH}  & 112 & 441&  32.1 & 39.8 & \textbf{43.1} \\\hline
entire  & 23.7k & 53.6k & 41.0 & 44.2 & \textbf{45.5}\\
\end{tabular}
\caption{Roost detection MAP for detector variants.\label{tab_exp_variants}}
\end{table}

\begin{figure*}
\centering
\includegraphics[width=\linewidth]{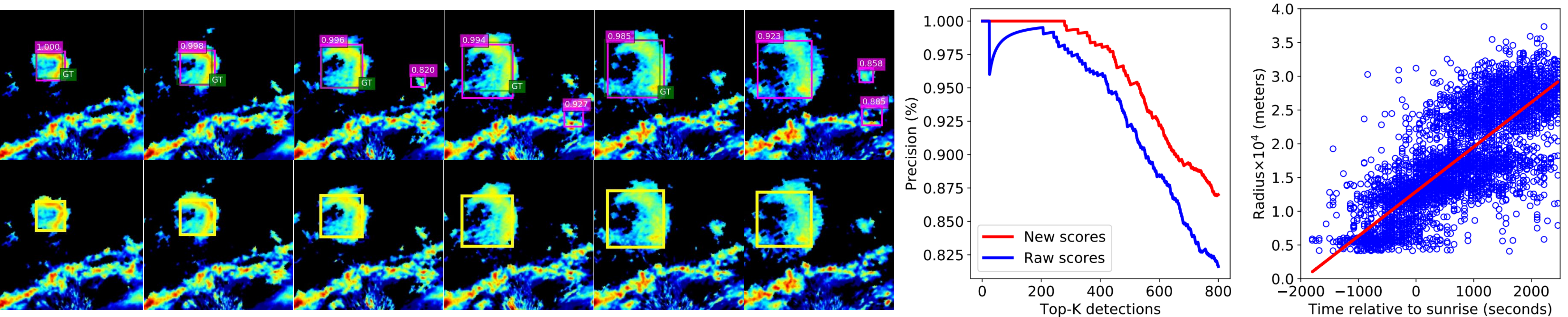}
\caption{Left: tracking example, with raw detections (top) and track (bottom). Transient false positives in several frames lead to poor tracks and are removed by the rescoring step. Middle: precision@k before and after rescoring. Right: Roost radius relative to time after sunrise.}
\label{fig:tracker}
\end{figure*}

\paragraph{Results: Tracking and Rescoring}
After obtaining the roost detections from our single-frame detector, we can apply our roost tracking model to establish roost tracks over time. Fig.~\ref{fig:tracker} shows an example radar sequence where roost detections have been successfully tracked over time and some false positives removed. We also systematically evaluated the tracking and rescoring model on scans from the \texttt{KOKX} station.
For this study we performed a manual evaluation of the top 800 detections before and after the contextual rescoring.
Manual evaluation was necessary due to human labeling biases, especially the omission of labels at the beginning or end of a roost sequence when roost signatures are not as obvious.
Fig.~\ref{fig:tracker}, middle panel, shows that the tracking and rescoring improves the precision across the entire range of~$k$. Our tracking model also enables us to study the roost dynamics over time (see Sec.~\ref{s:study} and Fig.~\ref{fig:tracker} right panel).

\section{Case study}\label{s:study}

\begin{figure*}
  \centering
  \begin{tabular}{c}
    \includegraphics[width = \linewidth]{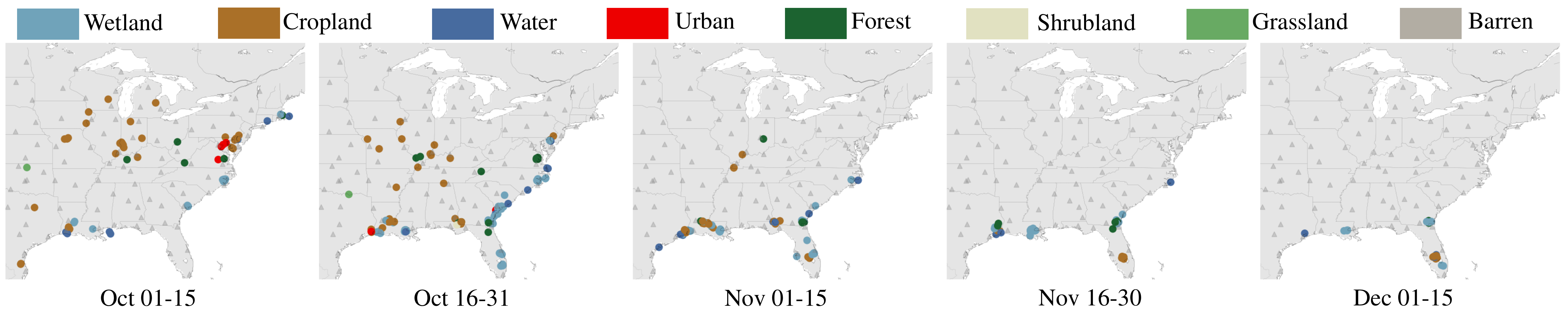} \\
  \end{tabular}
  \caption{Tree Swallow fall migration in 2013. The color circles show detected roost locations with each half-month period. The location of each roost is determined by the center of the first bounding box in the track, when the airborne birds are closest to their location on the ground. Faint gray triangles show radar station locations.}
  \label{fig:case-study}
\end{figure*}

We conducted a case study to use our detector and tracker to synthesize knowledge about continent-scale movement patterns of swallows.
We applied our pipeline to $419k$ radar scans collected from 86 radar stations in the Eastern US (see Figure~\ref{fig:case-study}) from October 2013 through March 2014. 
During these months, Tree Swallows are the only (fall/winter) or predominant (early spring) swallow species in the US and responsible for the vast majority of radar roost signatures.
This case study is therefore the first system to obtain comprehensive measurements of a single species of bird across its range on a daily basis. 
We ran our detector and tracking pipeline on all radar scans from 30 minutes before sunrise to 90 minutes after sunrise. We kept tracks having at least two detections with a detector score of 0.7 or more, and then ranked tracks by the sum of the detector score for each detection in the track.

\paragraph{Error Analysis} 
There were several specific phenomena that were frequently detected as false positives prior to post-processing.
We reviewed and classified all tracks with a total detection score of 5 or more prior to postprocessing (678 tracks total) to evaluate detector performance ``in the wild'' and the effectiveness of post-processing. This also served to vet the final data used in the biological analysis.
Tab.~\ref{table:errors} shows the number of detections by category before and after post-processing.
Roughly two-thirds of initial high scoring detections were swallow roosts, with another 5.6\% being communal roosts of \emph{some} bird species.

\begin{table}
  \begin{center}
    \begin{tabular}{|ccc|ccc|}
      \hline
      & Pre & Post &  & Pre & Post   \\
      \hline
      Swallow roost & 454 &  449 &       Other roost   &  38 &  38 \\
      Precipitation & 109 &  5 &        Clutter &  22 &  21 \\
      Wind farm     &  47 &  0  &       Unknown       &   8 &  8 \\
      \hline
    \end{tabular}
  \end{center}
  \caption{Detections by type pre- and post-filtering with auxiliary data. Post-processing effectively removes false positives due to precipitation and wind farms.}
  \label{table:errors}
\end{table}

\begin{figure}
\includegraphics[width= \linewidth]{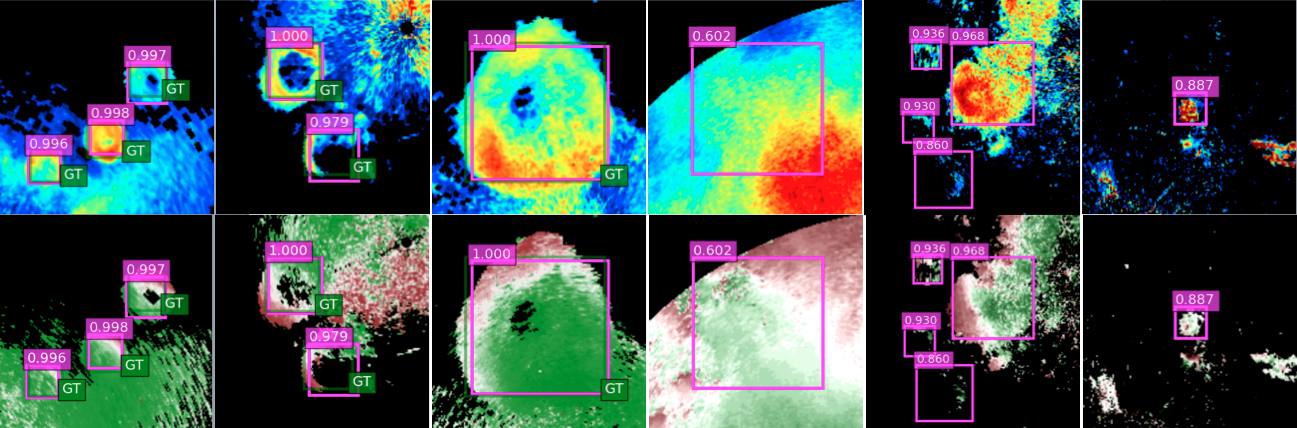}
\caption{\label{fig:detections} Some detections visualized on the reflectivity (top) and radial velocity (bottom) channels of different scans. The first three columns show swallow roost detections while the next three columns show detections due to rain, roosts of other species, and windmills.}
\end{figure}

The most false positives were due to precipitation, which appears as
highly complex and variable patterns in radar images, so it is common
to find small image patches that share the general shape and velocity
pattern of roosts (Fig.~\ref{fig:detections}, fourth column). Humans
recognize precipitation from larger-scale patterns and
movement. Filtering using $\rho_{HV}$ nearly eliminates rain false
positives.
The second leading source of false positives was wind farms. Surprisingly, these share several features of roosts: they appear as small high-reflectivity ``blobs'' and have a diverse velocity field due to spinning turbine blades (Fig.~\ref{fig:detections} last column). Humans can easily distinguish wind farms from roosts using temporal properties. \emph{All} wind farms are filtered successfully using the wind turbine database.
Since our case study focuses on Tree Swallows, we marked as ``other roost'' detections that were believed to be from other communally roosting species (e.g. American Robins, blackbirds, crows).
These appear in radar less frequently and with different appearance (usually ``blobs'' instead of ``rings''; Fig.~\ref{fig:detections}, fifth column) due to behavior differences.
Humans use appearance cues as well as habitat, region, and time of year to judge the likely species of a roost. We marked uncertain cases as ``other roost''.

\paragraph{Migration and Habitat Use} 
Fig.~\ref{fig:case-study} shows
swallow roost locations and habitat types for five half-month periods
starting in October to illustrate the migration patterns and seasonal
habitat use of Tree Swallows.
Habitat assignments are based on the majority habitat class from
the National Land Cover Database (NLCD)~\cite{usgs2011land}
within a \SI{10 x 10}{km} area surrounding the roost center, following
the approach of~\cite{bridge2016} for Purple
Martins. Unlike Purple Martins, the dominant habitat type
for Tree Swallows is wetlands (38\% of all roosts), followed by
croplands (29\%). 
These reflect the known habits of Tree Swallows to roost in reedy
vegetation---either natural wetlands (e.g. cattails and phragmites) or
agricultural fields (e.g., corn, sugar cane)~\cite{Winkler2011}.

In early October, Tree Swallows have left their breeding territories
and formed migratory and pre-migratory roosts throughout their
breeding range across the northern US~\cite{Winkler2011}.
Agricultural roosts are widespread in the upper midwest. 
Some birds have begun their southbound migration, which is evident by
the presence of roosts along the Gulf Coast, which is outside the
breeding range.
In late October, roosts concentrate along the eastern seaboard (mostly
wetland habitat) and in the central US (mostly cropland).
Most of the central US roosts occur near major rivers (e.g., the
Mississippi) or other water bodies.
The line of wetland roosts along the eastern seaboard likely
delineates a migration route followed by a large number of individuals
who make daily ``hops'' from roost to roost along this
route~\cite{winkler2006roosts}.
By early November, only a few roosts linger near major water bodies in
the central US.
Some birds have left the US entirely to points farther south, while
some remain in staging areas along the Gulf
Coast~\cite{Laughlin2016}.
By December, Gulf Coast activity has diminished, and roosts
concentrate more in Florida, where a population of Tree Swallows will
spend the entire winter.

Widespread statistics of roost locations and habitat usage throughout
a migratory season have not previously been documented, but are
enabled by our AI system to automatically detect and track roosts.
Our results are a starting point to better understand and
conserve these populations.
They highlight the importance of the eastern seaboard and Mississippi
valley as migration corridors, with different patterns of
habitat use (wetland vs. agricultural) in each.
The strong association with agricultural habitats during the harvest
season suggests interesting potential interactions between humans and
the migration strategy of swallows.

\paragraph{Roost Emergence Dynamics}
Our AI system also enables us to collect more detailed information
about roosts than previously possible, such as their dynamics over
time, to answer questions about their
behavior. Fig.~\ref{fig:tracker} shows the roost radius relative to
time after sunrise for roosts detected by our system.
Roosts appear around 1000 seconds before sunrise and expand at a
fairly consistent rate. The best fit line corresponds to swallows
dispersing from the center of the roost with an average airspeed
velocity of \SI{6.61}{m.s^{-1}} (unladen).

\section{Conclusion}\label{s:conclusion}
We presented a pipeline for detecting communal bird roosts using weather radar.
We showed that user-specific label noise is a significant hurdle to doing machine learning with the available data set, and presented a method to overcome this. 
Our approach reveals new insights into the continental-scale roosting behavior of migratory Tree Swallows, and can be built upon to conduct historical analysis using 20+ years of archived radar data to study their long-term population patterns in comparison with climate and land use change.

%\section*{Acknowledgments}
\paragraph{Acknowledgements}
This research was supported in part
by NSF \#1749833, \#1749854, \#1661259 and
the MassTech Collaborative for funding the UMass GPU
cluster.

\bibliographystyle{aaai}
{\small
\bibliography{roosts}
}

\end{document}